\documentclass[sn-vancouver,Numbered]{sn-jnl}

\usepackage{graphicx}%
\usepackage{multirow}%
\usepackage{amsmath,amssymb,amsfonts}%
\usepackage{amsthm}%
\usepackage{mathrsfs}%
\usepackage[title]{appendix}%
\usepackage{xcolor}%
\usepackage{textcomp}%
\usepackage{manyfoot}%
\usepackage{booktabs}%
\usepackage{algorithm}%
\usepackage{algorithmicx}%
\usepackage{algpseudocode}%
\usepackage{listings}%

\usepackage[export]{adjustbox}

\theoremstyle{thmstyleone}%
\theoremstyle{thmstyletwo}%

\theoremstyle{thmstylethree}%
\newtheorem{definition}{Definition}%

\begin{document}

\title[Differentially-Private Data Synthetisation]{Differentially-Private Data Synthetisation for Efficient Re-Identification Risk Control}


\author*[1]{\fnm{Tânia} \sur{Carvalho}}\email{tania.carvalho@fc.up.pt}

\author[2]{\fnm{Nuno} \sur{Moniz}}\email{nuno.moniz@nd.edu}

\author[1,3]{\fnm{Luís} \sur{Antunes}}\email{lfa@fc.up.pt}

\author[2]{\fnm{Nitesh} \sur{Chawla}}\email{nchawla@nd.edu}

\affil[1]{\orgname{Faculty of Sciences, University of Porto}, \orgaddress{\country{Portugal}}}

\affil[2]{\orgname{Lucy Family Institute for Data \& Society, University of Notre Dame}, \orgaddress{\state{Indiana}, \country{USA}}}

\affil[3]{\orgname{TekPrivacy}, \orgaddress{\city{Porto}, \country{Portugal}}}


\abstract{Protecting user data privacy can be achieved via many methods, from statistical transformations to generative models. However, all of them have critical drawbacks. For example, creating a transformed data set using traditional techniques is highly time-consuming. Also, recent deep learning-based solutions require significant computational resources in addition to long training phases, and differentially private-based solutions may undermine data utility. In this paper, we propose $\epsilon$-PrivateSMOTE, a technique designed for safeguarding against re-identification and linkage attacks, particularly addressing cases with a high \sloppy re-identification risk. Our proposal combines synthetic data generation via noise-induced interpolation with differential privacy principles to obfuscate high-risk cases. 
We demonstrate how $\epsilon$-PrivateSMOTE is capable of achieving competitive results in privacy risk and better predictive performance when compared to multiple traditional and state-of-the-art privacy-preservation methods, including generative adversarial networks, variational autoencoders, and differential privacy baselines. We also show how our method improves time requirements by at least a factor of 9 and is a resource-efficient solution that ensures high performance without specialised hardware.}

\keywords{Data Privacy, Synthetic Data, Deep Learning, Differential Privacy and Resource-Efficiency.}



\maketitle

\section{Introduction}\label{sec:intro}
Most traditional data de-identification tools use Privacy-Preserving Techniques (PPT) such as generalization (recoding values into broader categories) or suppression (replacing actual values with null values) to preserve the truthfulness of the data~\cite{carvalho2022survey}. Recent attention in data privacy has focused on data synthesis wherein deep learning models generate artificial data that reflect the statistical properties and structure of the original data set~\cite{el2020practical}. Although synthetic data generation is often perceived as a means to protect personal information fully, many questions have arisen on memorisation of deep learning models, especially outliers, which typically represent sensitive data points~\cite{nikolenko2019synthetic}. To address these privacy concerns, Differential Privacy (DP)~\cite{dwork2008differential} has been introduced in the synthetic data generation~\cite{torkzadehmahani2019dp}. 

These approaches are designed to protect an individual's privacy by preventing intruder attacks such as re-identification, also known as singling out (assigning a particular identity to an isolated record) or linkage attacks (linking at least two records concerning the same individual)~\cite{wp29}. 
However, a significant drawback is that the transformation of all instances may negatively affect data utility. As intruders typically target instances with the highest risk of exposing sensitive information~\cite{carvalho2021fundamental}, transforming all instances may inadvertently compromise the overall usefulness and effectiveness of the data. Most importantly, current privacy-preserving approaches still exhibit significant computational and time overheads~\cite{bird2020reducing}.


In this paper, we propose $\epsilon$-PrivateSMOTE, a new strategy for privacy-preserving tabular data sharing using synthetic data generation. 
Our focus lies on addressing the re-identification risk by strategically replacing high-risk cases with synthetic and similar data points. 
To our knowledge, interpolation methods have not yet been thoroughly explored in favour of privacy. 
In this context, we encounter exciting opportunities to explore existing strategies.
Our contribution involves leveraging SMOTE-inspired heuristics~\cite{chawla2002smote}, a sophisticated interpolation technique based on nearest neighbours algorithms, that can operate effectively even in scenarios with limited data quantities~\cite{pradipta2020improving}.
Also, it enables targeted enhancements of privacy assurances by focusing on the highest-risk cases. 
Our approach takes advantage of the stochastic indistinguishability of DP, which aims to protect against several privacy attacks (e.g., differencing, linkage, and reconstruction attacks)~\cite{dwork2014algorithmic} using a privacy budget parameterized by $\epsilon$. 
The incorporation of the Laplace mechanism, a well-established differentially private mechanism~\cite{prooflaplace}, ensures statistically indistinguishable analyses on similar datasets.

While current privacy-preserving solutions for tabular data sharing were designed to modify all instances,   
$\epsilon$-PrivateSMOTE distinguishes for its ability to strategically enhance privacy assurances in a targeted way offering a promising avenue for secure data sharing and simultaneously maximising data utility.
The significance and relevance of our proposal can be summarised as follows:

\begin{itemize}
    \item $\epsilon$-PrivateSMOTE is easy to implement and configure when compared to traditional privacy solutions using PPT (data granularity reduction);
    \item $\epsilon$-PrivateSMOTE outperforms deep learning- and differentially private-based solutions in predictive performance;
    \item $\epsilon$-PrivateSMOTE produces more new data variants with fewer resources than deep learning- and differentially private-based approaches, both in time and computation;
    \item Contrary to any other privacy-preserving solution tested, $\epsilon$-PrivateSMOTE offers a notable balance between privacy and predictive performance.
\end{itemize}

The remainder of the paper is organized as follows. Section~\ref{sec:related_work} overviews related work on strategies for privacy-preserving data sharing and sampling strategies. We present $\epsilon$-PrivateSMOTE approach in Section~\ref{sec:smote}. An extensive experimental evaluation is described in Section~\ref{sec:expstudy}. A discussion of the results obtained is introduced in Section~\ref{sec:discussion}, and conclusions are provided in Section~\ref{sec:conclusion}.

\section{Related Work}\label{sec:related_work}


In the de-identification process, it is fundamental to make assumptions about the intruder's background knowledge since it is unknown what information the person may possess. Such assumptions are based on the selection of quasi-identifiers (QI): attributes that, when combined, can lead to identification, e.g., date of birth, gender, and profession. 
To limit the disclosure risk of personal information, it is imperative to apply appropriate transformations on such QI~\cite{domingo2008survey,carvalho2022survey}.

The transformed data set is then evaluated for its re-identification risk. Two standard measures are $k$-anonymity and record linkage. $K$-anonymity~\cite{samarati2001protecting} indicates how many $k$ occurrences occur in the data set w.r.t a specific combination of QI values. An intruder can single out a record when $k=1$. On the other hand, record linkage (or linkability)~\cite{anonymeter} aims to measure the ability of re-identification by linking two records using similarity functions. 
Privacy is also a focus during the machine learning training phase; Membership Inference (MI) risk aims to identify whether a given data point is included in the training set~\cite{long2017towards}.

\subsection{Strategies for De-identification}

De-identified data is achieved via distinct approaches. We analyse the three most common: traditional, deep learning- and differentially private-based solutions.

\textbf{Traditional techniques} uses PPT to achieve a certain level of privacy. These may be validated using various tools, such as $k$-anonymity~\cite{samarati2001protecting} or Differential Privacy (DP)~\cite{dwork2008differential}. 
To achieve $k$-anonymity, each individual represented in a data set cannot be distinguished from at least $k - 1$ other individuals. Data is typically reduced, as is its granularity, which is likely to impact the predictive performance of machine learning models~\cite{brickell2008cost}. 
On the other hand, to achieve DP, the statistical results of the data set are not supposed to be influenced by an individual's contribution. Although DP has gained prominence recently, it is often applied to queries or during the training process in machine learning applications. Nevertheless, a few approaches have emerged for releasing differentially private data sets; for instance, ($k$, $\epsilon$)-anonymity~\cite{holohan2017k} uses DP on a given attribute. However, their proposal must first guarantee a certain level of $k$-anonymity and then apply the noise to each group w.r.t QI (equivalence classes). Moreover, this technique is applied to a single attribute. 
Recently, Muralidhar et al.~\cite{muralidhar2020epsilon} showed that a fixed $\epsilon$ does not guarantee a certain level of confidentiality or utility on differentially-private data created using noise-added covariances approach nor using added noise to the cumulative distribution function.

\textbf{Deep Learning-based solutions}
such as CTGAN~\cite{xu2019modeling} emerged as an alternative to traditional techniques. A comparison of generative methods was performed with particular attention to privacy concerns. This comparison shows that medical data generated by GANs does not model outliers well and that such models are also vulnerable to MI attacks~\cite{yale2019assessing}.

\textbf{Differentially Private-based solutions}
incorporate DP into the process of synthetic data generation for high privacy guarantees. For instance, DPGAN~\cite{xie2018differentially}, DP-CGAN~\cite{torkzadehmahani2019dp} and PATE-GAN~\cite{jordon2018pate} are differentially private GAN models that add noise during the training process of the discriminator. 
DP has also been added to the tree-based machine learning models to generate synthetic data~\cite{mahiou2022dpart}. Besides, PrivBayes~\cite{zhang2017privbayes} uses DP in a Bayesian Network focusing on marginal distributions. Recent examples on marginals include PrivSyn~\cite{privsyn} and PrivMRF~\cite{privmrf}.

\subsection{Sampling Methods}

Although privacy-preserving data sharing focuses on data transformations or generative solutions, we suggest that sampling methods, commonly used in imbalanced learning tasks, could be a valuable tool to tackle this challenge. 

Popular data pre-processing (i.e., sampling) methods in imbalanced classification tasks include random under- (RUS) and over-sampling (ROS). The former randomly selects cases of the majority class for removal 
while the latter works by replicating cases of the minority class. 
The most popular is SMOTE (Synthetic Minority Over-sampling Technique)~\cite{chawla2002smote}, an over-sampling strategy that focuses on the synthetic generation of rare cases by interpolating existing ones, based on the nearest neighbour algorithm. 
BorderlineSMOTE~\cite{han2005borderline} is a variation of SMOTE that focuses on oversampling minority examples near class borderlines. In addition to these approaches, several existing sampling methods have been reviewed to address class imbalance problems~\cite{tarekegn2021review, spelmen2018review, survey_imb_22}.



\subsection{Summary}
The main challenge in the de-identification process using PPT is to achieve an appropriate level of privacy and its utility, as it involves several steps until the balance between the two metrics is found. Furthermore, a de-identification level is stipulated, e.g., in the case of $k$-anonymity, an algorithm ensures that the data respects this $k$ level. 
However, if the algorithm cannot use more generalisation, much information is lost as suppression is applied. In addition, data points that initially would not have a high risk of re-identification are altered, affecting the utility. On the other hand, adding noise through DP can potentially result in data with an increased level of uncertainty, which can be difficult for data analysts to work with and can lead to inaccurate machine learning models.

Finally, a significant drawback of generative models using deep learning solutions is that statistical properties may not be well captured with smaller data sets. Moreover, several implementations of such tools have been shown to violate privacy guarantees, exposing many records to inference attacks and failing to retain data utility~\cite{stadler2022synthetic}. Additionally, a major challenge of deep learning-based solutions is their complexity, as they are computationally costly in memory and require significant execution time~\cite{bird2020reducing}. 

Despite increased efforts to improve de-identification strategies, the above limitations motivate us to explore efficient privacy alternatives. An underinvested area is the intersection with sampling methods used in imbalanced learning tasks, such as SMOTE~\cite{chawla2002smote}. In contrast to deep learning-based solutions, the heuristics of interpolation methods are simple, allowing synthetic data to be generated with less effort. We propose to explore such an approach for data privacy.

\section{Privacy-Preserving Data Synthetisation}\label{sec:smote}

In this section, we present a novel approach for privacy-preservation leveraging the indistinguishability of DP and interpolation methods for synthetic data generation, focusing on cases with a high re-identification risk.

Consider a data set $T = \{t_1, ..., t_n\}$, where $t_i$ corresponds to a tuple of attribute values for an individual's record. Let  $V = \{v_1, ..., v_m\}$ be the set of $m$ attributes. 
A set of quasi-identifiers (QI) consists of attributes values that could be known to the intruder for a given individual where $QI \in V$ and is denoted as $QI$ = $\{v_{j_{1}}, ..., v_{j_{k}}\}$. 
For example, $QI = \{v_{j_1}, v_{j_2}, v_{j_3}\}$ might represent an individual's gender, age, and zip code.
A tuple of selected QI is denoted as 
$t^{QI}_i = \{t_{i,v_{j}} : v_j \in QI\}$. 
\newline

\begin{definition}[\textbf{Re-identification}]
Assume an intruder knows that an individual with the tuple $t^{QI}_r$ is in $T$. If no other tuple shares identical information, then $t^{QI}_r$ is singled out and the intruder can uniquely identify the individual. 
\newline
\end{definition}

Additionally, an intruder could infer that there is exactly one person in $T$ who has $t^{QI}_u \langle$gender: male, age: 75, zip: 46711$\rangle$. While singling out this particular individual may not constitute re-identification, the ability to isolate an individual may lead to other privacy attacks~\cite{anonymeter}. Therefore, it is essential to increase safeguarding measures. For example, if another $t^{QI}_p$ shares the same set of information as $t^{QI}_r$, there is a 50\% probability of re-identification. This reinforces the importance of carefully examining other tuples that may share similar QI. According to the $k$-anonymity definition, to ensure the protection of cases with a 50\% risk of re-identification, the protected version of $T$ must be at least 3-anonymous. This implies that each group w.r.t. QI contains at least three indistinguishable individuals, reducing the probability of re-identification to 33\% for each individual. Such criteria can be adjusted by the $k$ parameter according to the use case under analysis. In our study, we focus on protecting single outs and cases with a 50\% risk of re-identification. The $k$-anonymity basis is used to detect and mitigate the risk associated with these highest-risk instances.
\newline
\begin{definition}[\textbf{Highest-risk selection}]
Given $T$ and $T[QI]$, $T'$ corresponds to the highest-risk cases table, where each sequence of values in a tuple $t_{i,v_j}$ occurs at a maximum two times in $T[QI]$. \newline
\end{definition}

As such, instead of generating synthetic instances via interpolation of cases from the minority class, we focus on the highest-risk instances. For such instances, we generate numerical synthetic cases based on randomly weighted interpolation of nearest neighbours, akin to the process in SMOTE~\cite{chawla2002smote}, without any class discrimination, i.e., focusing solely on case proximity. 
For the interpolation of nominal attributes, we employ a random selection process where a category is chosen from the neighbours of each instance. However, in cases where only one category is available for the nearest neighbours, the selection is made based on the categories observed in the entire sample to prevent the new value from being equal to the original. 

At this point, we still face a major drawback: the inability to deal with lower data variability. 
In some cases, the nearest neighbour chosen for synthesis may be equal to the instance to be synthesised (e.g. \textit{Age} attribute), generating the same value as the original, which is critical for privacy. To overcome such a drawback, we propose to enforce privacy by forcing a difference between the two values using the standard deviation (\textit{std}), i.e. multiplying the random uniform value [0,1] by $std$ or $-std$. 

Nevertheless, we enhance privacy guarantees through DP by replacing the standard uniform random selection with a Laplace-based mechanism that introduces noise from the Laplace distribution. A randomised algorithm $M$ satisfies $\epsilon$-differential privacy~\cite{dwork2008differential} if for all data sets $D_1$ and $D_2$ differing on at most one element, which differ in a single individual, and for all subsets of possible outcomes S ($S \subseteq Range(M)$),
\begin{equation}
    Pr[M(D_1) \in S] \leq e^\epsilon \times Pr[M(D_2) \in S]
\end{equation}

Generally, uniform noise is (0,$\delta$)-differential private~\cite{he2020differential}, where $\epsilon = 0$ means that no information is revealed about any individual's data, also known as ``pure'' differential privacy; in opposite, data utility is destroyed. The $epsilon$ parameter indicates the amount of added noise, often referred to as \textit{privacy budget} or \textit{cost} and $\delta$ represents the probability of violating the privacy cost~\cite{dwork2008differential}.
Therefore, we strengthen privacy protection by leveraging the Laplace mechanism, which is differentially private~\cite{prooflaplace}.
In our context, this mechanism is described as follows. 
\newline
\begin{definition}[\textbf{Laplace mechanism}]
For each value $\vartheta$ in a tuple $t_{i,v_j}$ of $T'[QI]$, the Laplace mechanism $A$ is given by  $A = \vartheta + Laplace(1/\epsilon)$. \newline
\end{definition}

We call this proposed solution $\epsilon$-PrivateSMOTE. For clarity and further reproducibility, we provide a detailed description of the execution in Algorithm~\ref{alg:PrivateSMOTE}. The Algorithm is organised into distinct steps, which are outlined below.

\begin{enumerate}
    \item nearest neighbours are computed over the entire sample (line 10);
    \item $k$ highest-risk cases are determined based on the selected QI (line 11);
    \item $knn$ nearest neighbours are found for each highest-risk instance (line 13);
    \item DP is incorporated instead of using uniform noise as a privacy protection mechanism (lines 18-19);
    \item for stronger privacy guarantees, the synthesis of a new case is enforced when a specific example is equal to the chosen nearest neighbour (line 19 \#2) and,
    \item nominal attributes are interpolated by randomising the selection based on the uniqueness of a case's neighbours. If there is no randomness in the former case, the uniqueness of the entire sample is considered (lines 20-21).
    
\end{enumerate}

\begin{algorithm}[ht!]
\scriptsize
\caption{$\epsilon$-PrivateSMOTE}\label{alg:PrivateSMOTE}
\begin{flushleft}
\textbf{Input:} Table $T$ with set of $m$ attributes and $n$ instances, amount of new cases $N$, number of nearest neighbours $knn$, amount of noise $\epsilon$, group size $k$ and QI set. \\
\textbf{Output:} $N * n$ synthetic samples
\end{flushleft}
\begin{algorithmic}[1]
\Function{HighestRisk}{$T, k, \text{QI}$}
    \State $eq$ = $\text{compute the occurrences of each group in $T$ w.r.t QI}$
    \State $\text{$T'$ = filter rows where $eq$ is less than $k$}$
    \State \textbf{return} $T'$
\EndFunction

\Procedure{$\epsilon$-PrivateSMOTE}{$T, N, knn, \epsilon, k, \text{QI}$}

    \State $\text{$y$ = class examples $(y_{1},..., y_{n})$ in }T$
    \State $\text{synthetic\_samples} = []$
    \State $\zeta\text{ = one-hot-encoding and standardisation of }T$ 
    \State $nn\text{ = fitted nearest neighbours using }\zeta$
    \State $T'$ = \Call{HighestRisk}{$T, k, \text{QI}$} 
    \For{$i \gets 1$ to $n$ instances}
            \State $nnarray\text{ = $knn$ nearest neighbours from $nn$ if $i \in$ indexes of } T'$
\While{$N \neq 0$ (populate $N$ times)}
            \For{$attr \gets 1$ to $m$}
                \State $a\text{ = instance in position }i$
                \State $b\text{ = random instance from }nnaray$

                    \If{$T[attr] \text{is numerical}$}
                     \State $\text{new\_instance} =$
                     \begin{minipage}[t]{\linewidth}
                         $ a + Laplace(0,1/\epsilon) * (b-a) \textbf{ if } a \neq b \text{\hspace{6em} $\#1$} $ \\ 
                          $\textbf{ else } a + Laplace(0,1/\epsilon) * rand[std,-std] \text{ \hspace{3em} \#2}$ 
                        \end{minipage}
                        
                    \ElsIf{$T[attr] \text{is nominal}$}
                    \State $\text{new\_instance} =$
                    \begin{minipage}[t]{\linewidth}
                        $\text{random category from } T[nnarray] \text{ unique values}$ \\
                         $\text{if there are two or more categories,}$ \\
                         $\text{otherwise from } T[attr] \text{ unique values}$
                        \end{minipage}
                    \EndIf
                    \State $\text{add new\_instance to synthetic\_samples}$
                \EndFor
                \State $\text{add the intact target value $y[i]$ to synthetic\_samples}$

            \EndWhile
    \EndFor
\EndProcedure
\end{algorithmic}
\end{algorithm}

\subsection{Threat model}
Although synthetic data is at the forefront of data de-identification methods, unresolved privacy concerns still require attention and mitigation. Related literature on differentially private-based solutions essentially focuses on presenting privacy assurances concerning different levels of DP~\cite{xie2018differentially,jordon2018pate,privsyn} and often lacks comprehensive discussions on associated threat models. In addition, deep learning- and differentially private-based models are generally
evaluated against MI attacks~\cite{domias}. However, this type of attack is conventionally applied to learning models~\cite{domias}, specifically during the training phase, making MI attacks less applicable to traditional techniques and $\epsilon$-PrivateSMOTE. On the one hand, in a $k$-anonymised data set, each record is deliberately made to resemble multiple other records, making it challenging to determine the presence of a particular individual's data point; on the other hand, as the core of our proposal lies in the heuristics of the interpolation method, the augmentation and DP noise makes it challenging to determine whether a data point corresponds to a specific individual in the original data set. Therefore, membership attacks are not as informative for assessing the privacy of data sets using these approaches.

Linkability~\cite{giomi2022unified} is a common privacy attack that involves a concerted effort by an unauthorised entity to establish a correlation between two records~\cite{carvalho2021fundamental}. The intruder attempts to link and potentially expose sensitive information in the targeted records, with outliers being particularly vulnerable. 
The target records of the attack are a collection of $N_A$ original records randomly drawn from the original data set.
Suppose an intruder has some knowledge of the targets and access to two data tables, a synthetic data table ($T_A$) and another obtained from public information ($T_B$), both of which contain these targets. The intruder's goal is to correctly match records between $T_A$ and $T_B$ given a set of QI common to both data sets. 
\newline
\begin{definition}[\textbf{Linkability~\cite{giomi2022unified}}]
For each instance in $T_A$, an intruder finds the $k$ closest synthetic instance in $T_B$. The resulting indices are $I^A= (I_i^A,...,I_{N_A}^A )$, where each $I_i^A$ is the set of $k$ indexes of the synthetic
records that are nearest neighbours of the $i^{th}$ target in the subspace of feature set.
\newline
\end{definition}

For each of the $N_A$ targets, it is checked whether both nearest neighbours sets share the same synthetic data record. In a positive case, an attacker can associate previous unconnected information about a target individual that is present in the original data set. Each successfully established link is scored as a success, where the outcome ($O$) is determined by:

\begin{equation}
   O_i(I_i^A,I_i^B)=\begin{cases}
      1, & \text{if $I_i^A \cap I_i^B$ $\neq$ 0}\\
      0, & \text{otherwise.}
    \end{cases} 
\end{equation}

\section{Experimental Evaluation}\label{sec:expstudy}

In this section, we provide a thorough experimental evaluation. Our goal is to answer the following research questions.  
How does $\epsilon$-PrivateSMOTE compare to competing methods in terms of predictive performance and privacy risk, namely linkability (\textbf{RQ1}), how do optimisation paths (prioritisation of a single vector) affect the results of both vectors (\textbf{RQ2}), and how does $\epsilon$-PrivateSMOTE behave in terms of time/computational resources complexity (\textbf{RQ3})? In the following, we present our methodology and experimental evaluation methods. Then, we briefly describe the data we used. Finally, we present the results of the experiments.

\subsection{Methodology and Methods}\label{subsec:methodology}

We divide our methodology into three phases: \textit{i)} data transformation, \textit{ii)} data utility concerning general distributions and predictive performance evaluation, and \textit{iii)} data privacy. Regarding data transformation (\textit{i}), we apply five approaches to original data sets. For each set, we generate several variants employing the following methodologies.

\textbf{Traditional Techniques} are applied according to the characteristics of the QI, i.e., the type of attribute values. We assume that any attribute can be a QI as we do not know what information an intruder might possess. Consequently, we collect five different QI sets, each consisting of a random selection of 40\% of the attributes, that represent potential background knowledge for an intruder. 
For each set, we create new transformed variants. We use the ARX tool~\cite{prasser2020flexible}, a benchmark solution. We apply generalisation with specific hierarchies using specific intervals. We also select the outlier suppression option to achieve the desired privacy level. From the solution space, we select the three best solutions w.r.t utility (information loss) as provided by the tool.

\textbf{Sampling methods} are applied to all attributes and instances. 
We use RUS, SMOTE, and BorderlineSMOTE as implemented in the Python library \textit{imblearn}~\cite{imblearn}.  
Regarding parameterization, we use $ratio \in \{0.25, 0.5, 0.75, 1\}$ for RUS, and $ratio \in \{0.5, 0.75, 1\}$ and $knn \in \{1, 3, 5\}$ for SMOTE and Borderline. 


\textbf{Deep Learning-based solutions}
are also applied to all attributes and instances. We use the \textit{SDV} library~\cite{sdv}, experiment with Copula GAN, TVAE, and CTGAN and the parameters: $epochs$ $\in$ $\{100, 200\}$ and $batch\_size$ $\in$ $\{50, 100\}$. 

\textbf{Differentially Private-based solutions} are still in an improvement phase\footnote{We attempted to use PrivBayes; however, we encountered challenges in generating data variants. The program exhibited long execution times, using only the CPU.}. 
We use \textit{synthcity}~\cite{synthcity} that provides a benchmark for synthetic data generators. Our experimentation involved the use of DPGAN and PATE-GAN with the same $epochs$ $\in$ $\{100, 200\}$, $batch\_size$ $\in$ $\{50, 100\}$ and $\epsilon \in \{0.1, 0.5, 1.0, 5.0\}$.

\textbf{$\epsilon$-PrivateSMOTE} is applied to the highest-risk cases given the set of QI (the same as for traditional techniques). We use $N \in \{1, 2, 3\}$ (number of times a case is repeated) and $knn \in \{1, 3, 5\}$. Also, we need to specify the amount of added noise. We experiment with \sloppy $\epsilon \in \{0.1, 0.5, 1.0, 5.0, 10.0\}$. Previous solutions do not include $\epsilon = 10.0$ as it fails to generate several data variants. 
\newline

All privacy-preserving data variants are evaluated for data utility (\textit{ii}). For general utility, we use SDMetrics from SDV~\cite{sdv}. Specifically, we use Range Coverage, 
Boundary Adherence, 
Statistical Similarity 
and Correlation 
to check the extent to which each data variant differs from the original data in terms of general distributions. 

We also evaluate the predictive performance using the following methodology. We apply four classification algorithms (Random Forest~\cite{ho1998random}, XGBoost~\cite{chen2016xgboost}, Logistic Regression~\cite{logit} and Neural Network~\cite{NN}) using \textit{Scikit-learn}~\cite{pedregosa2011scikit}, to test all the privacy strategies. Final models for each algorithm are chosen based on a 2*5-fold cross-validation estimation of the evaluation scores for models based on a grid search method. We also obtain performance results on test data -- 20\% of the original data set. Table~\ref{tab:algorithms} summarises this information. 

\begin{table}[!ht]
\begin{center}
    \scriptsize
    \begin{adjustbox}{max width=0.75\linewidth}
\begin{tabular}{@{}l|l@{}}
\toprule
\textbf{Algorithm}  & \textbf{Parameters}                                       \\ \midrule
Random Forest       & \begin{tabular}[c]{@{}l@{}}$n\_estimators \in \{100, 250, 500\}$\\ $max\_depth \in \{4, 7, 10\}$\end{tabular}                                                                                       \\ \midrule
Boosting            & \begin{tabular}[c]{@{}l@{}}$n\_estimators \in \{100, 250, 500\}$\\ $max\_depth \in \{4, 7, 10\}$\\ $learning\_rate \in \{0.1, 0.01\}$\end{tabular}                                                                     \\ \midrule
Logistic Regression & \begin{tabular}[c]{@{}l@{}}$C \in \{0.001, 1, 10000\}$\\ $max\_iter \in \{10e^5, 10e^6\}$\end{tabular}                                                               \\ \midrule
Neural Network & \begin{tabular}[c]{@{}l@{}}$hidden\_layer\_sizes \in$ \{{[}$n\_feat${]}, {[}$n\_feat / 2${]}, {[}$n\_feat * 2/3${]},\\                                      {[}$n\_feat, n\_feat / 2${]}, {[}$n\_feat, n\_feat * 2/3${]}, {[}$n\_feat / 2, n\_feat *  2/3${]},\\                                      {[}$n\_feat, n\_feat  / 2, n\_feat  * 2/3${]}\}\\ $alpha \in \{0.005, 0.001,  0.0001\}$\\ $max\_iter \in \{10000, 100000\}$\end{tabular}  \\
\bottomrule
\end{tabular}
\end{adjustbox}
\end{center}    
\caption{Learning algorithms considered in the experimental evaluation and respective hyperparameter grid. Each line of parameter \textit{hidden layer sizes} (Neural Networks) represents an increasing number of layers (1--3).}
    \label{tab:algorithms}
\end{table}

The effectiveness of predictive performance will be evaluated using AUC (Area Under the ROC Curve)~\cite{weng2008new}.
We also apply statistical tests using Bayes Sign Test~\cite{Benavoli2017} to evaluate the significance of our experimental results. We use the percentage difference between each pair of solutions as $\frac{R_a - R_b}{R_b} * 100$ where $R_a$ is the solution under comparison and $R_b$ is the baseline solution. In this context, ROPE (Region of Practical Equivalence)~\cite{kruschke2015bayesian} is used to specify an interval ([-1\%, 1\%]) where the percentage difference is considered equivalent to the null value. 
Thus, \textit{1)} if the percentage difference of the specified metric between solutions \textit{a} and \textit{b} (baseline) is greater than 1\%, solution \textit{a} outperforms \textit{b} (win); \textit{2)} if the percentage difference is within the specified range, they are of practical equivalence (draw); and \textit{3)} if the percentage difference is less than -1\%, the baseline outperforms solution \textit{a} (lose).

Each resulting data variant is also evaluated regarding privacy risk (\textit{iii}). We evaluate privacy through linkability risk by comparing each variant against the original data set. The comparison space includes only the specified set of QI~\footnote{For comparison purposes, we use the same set of QI for sampling and deep learning- and differentially private-based solutions.}. We use \textit{Anonymeter} library~\cite{anonymeter} for linkability assessment. 
For this evaluation, it is required a control data set which we use the test data set (20\% of original data), and $k$ that indicates the search space for nearest neighbours; we use $k=10$. 

\begin{figure}[ht!]
   \centering
   \includegraphics[width=0.75\linewidth]{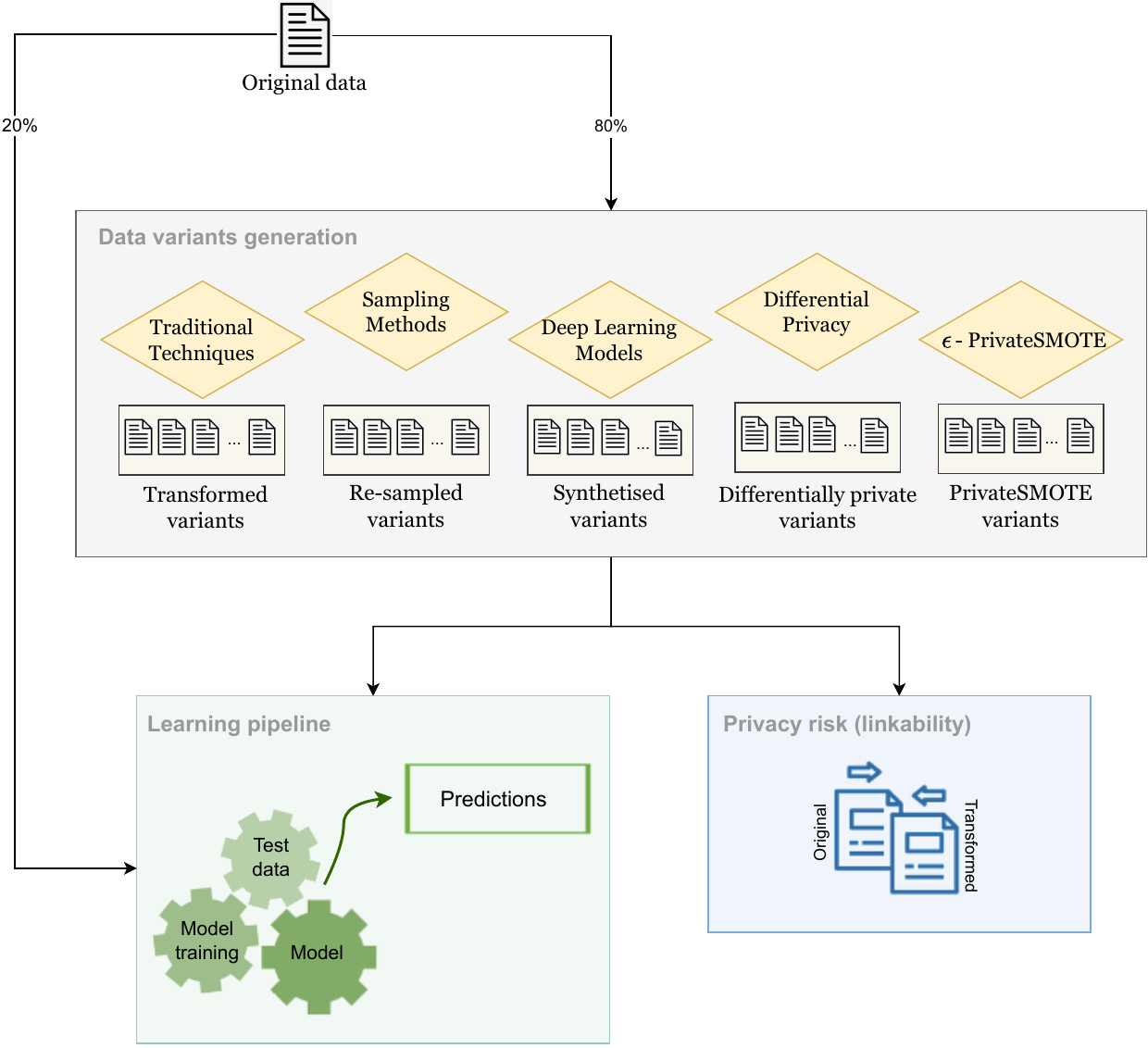}
 \caption{Methodology of the experimental evaluation.}
 \label{fig:workflow}
\end{figure}

For conciseness, our experimental methodology is illustrated in Figure~\ref{fig:workflow}.
 
\subsection{Data}\label{subsec:data}
To provide an extensive experimental study, we collected 15 diverse classification data sets from the public and open data repository OpenML~\cite{OpenML2013}. These datasets span various domains, ensuring different representations of real-world scenarios. Common characteristics include:
\textit{i)} binary class target, \textit{ii)} number of instances between 1.043 and 15.545, and \textit{iii)} number of features greater than five and less than 103. The average number of instances of the retrieved data sets is 4.916. Also, there is an average of 28 numerical and two nominal attributes. For transparency and reproducibility, all datasets used in our experimental evaluation are publicly available \footnote{https://www.kaggle.com/datasets/up201204722/original}.

\subsection{Results}\label{subsec:results}

The first set of results from the experimental evaluation focuses on an overall assessment of data utility and privacy risk for $\epsilon$-PrivateSMOTE.
Concerning predictive performance, results refer to the out-of-sample performance of the best models for each data variant, estimated via cross-validation. Then, we calculate the percentage difference between the best-estimated model for each data variant solution and the baseline -- best solution for each original data set.

\begin{figure}[ht!]
   \centering
   \scriptsize
   \includegraphics[width=0.65\linewidth]{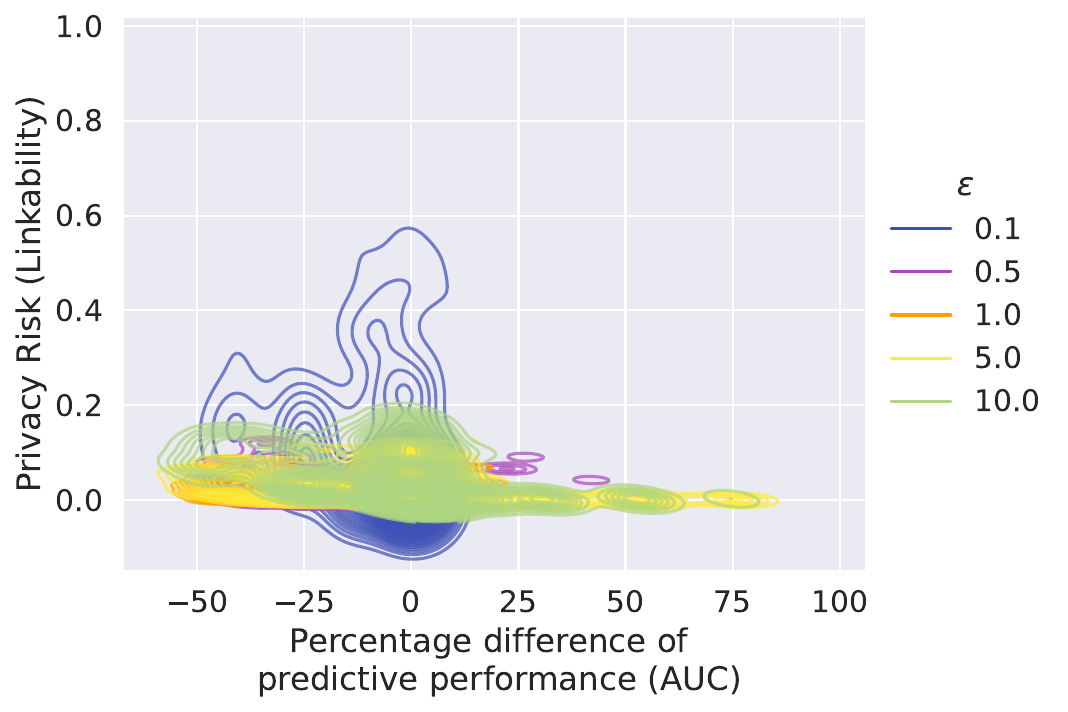}
 \caption{Relationship between privacy risk and predictive performance of all data variants produced with $\epsilon$-PrivateSMOTE for each $\epsilon$.}
 \label{fig:density}
\end{figure}

Figure~\ref{fig:density} illustrates the correlation between the linkability and the percentage difference of predictive performance for each $\epsilon$ across all produced data variants.

\begin{figure}[ht!]
   \centering
   \scriptsize
   \includegraphics[width=\linewidth]{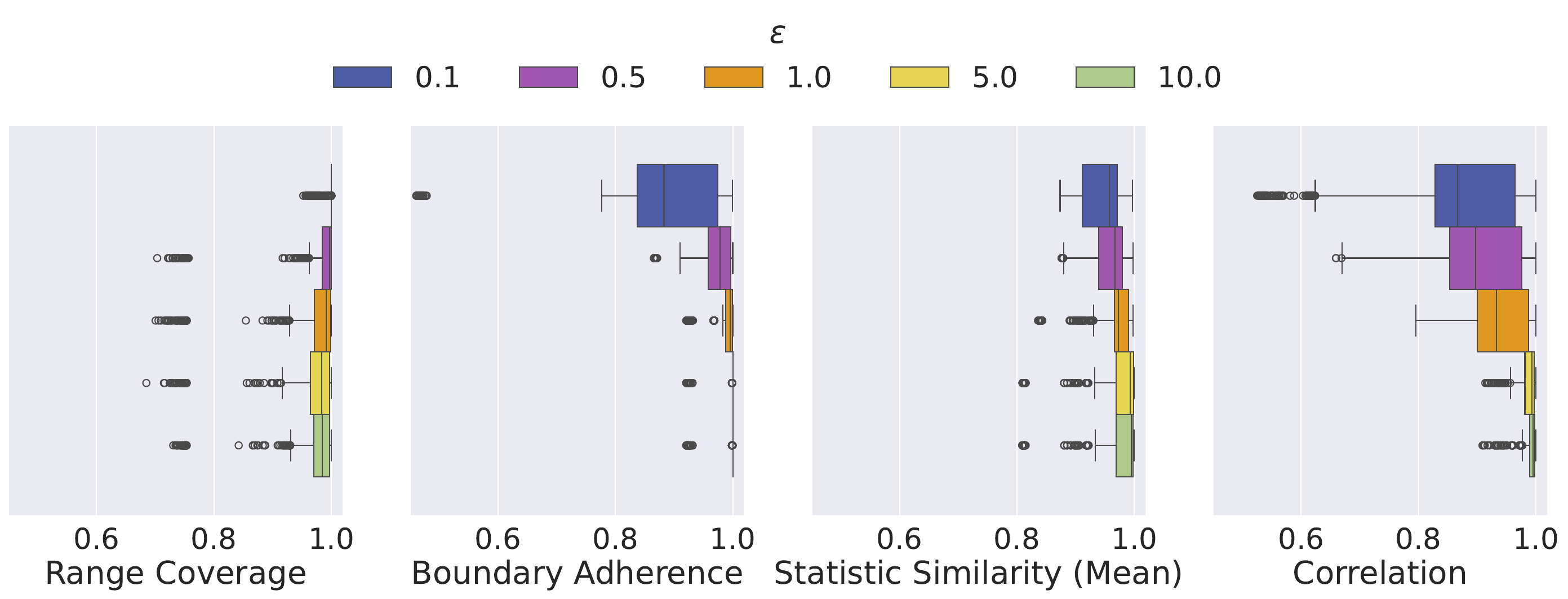}
 \caption{Data utility measures for each $\epsilon$ across all data variants using $\epsilon$-PrivateSMOTE.}
 \label{fig:utilep}
\end{figure}

Overall, we observe a central tendency around zero for both the percentage difference of predictive performance and linkability. Nevertheless, higher $\epsilon$ values can provide solutions with high predictive performance and lower privacy risk. Such a result suggests that $\epsilon$-PrivateSMOTE is able to provide solutions that maximise both vectors, in particular, for $\epsilon \geq 0.5$. Although it is well-known that lower $\epsilon$ values, i.e. higher noise, result in higher privacy~\cite{dwork2008differential} and consequently may negatively affect the predictive performance, we observe that $0.1$-PrivateSMOTE has higher linkability and yet can produce several solutions with equivalent predictive performance. This outcome is potentially ascribed to the larger values at the tails of the distribution, which may lead to an increase in the overall distribution, as we can evidence in Figure~\ref{fig:utilep}.


This Figure shows different utility measures concerning all $\epsilon$ values for all data variants created with $\epsilon$-PrivateSMOTE. We conclude that lower $\epsilon$ values present better Range Coverage meaning that the data variants cover the full range of each numeric column.
On the other hand, lower $\epsilon$ fails to adhere to the real boundaries. Such conditions result in more spread values at the tails, facilitating linkability since these values are rarer.
Also, it provides lower Statistic Similarity and Correlation compared to high $\epsilon$ values. In general, we observe better utility for higher $\epsilon$ values.

\begin{figure}[ht!]
   \centering
   \scriptsize
   \includegraphics[width=0.9\linewidth]{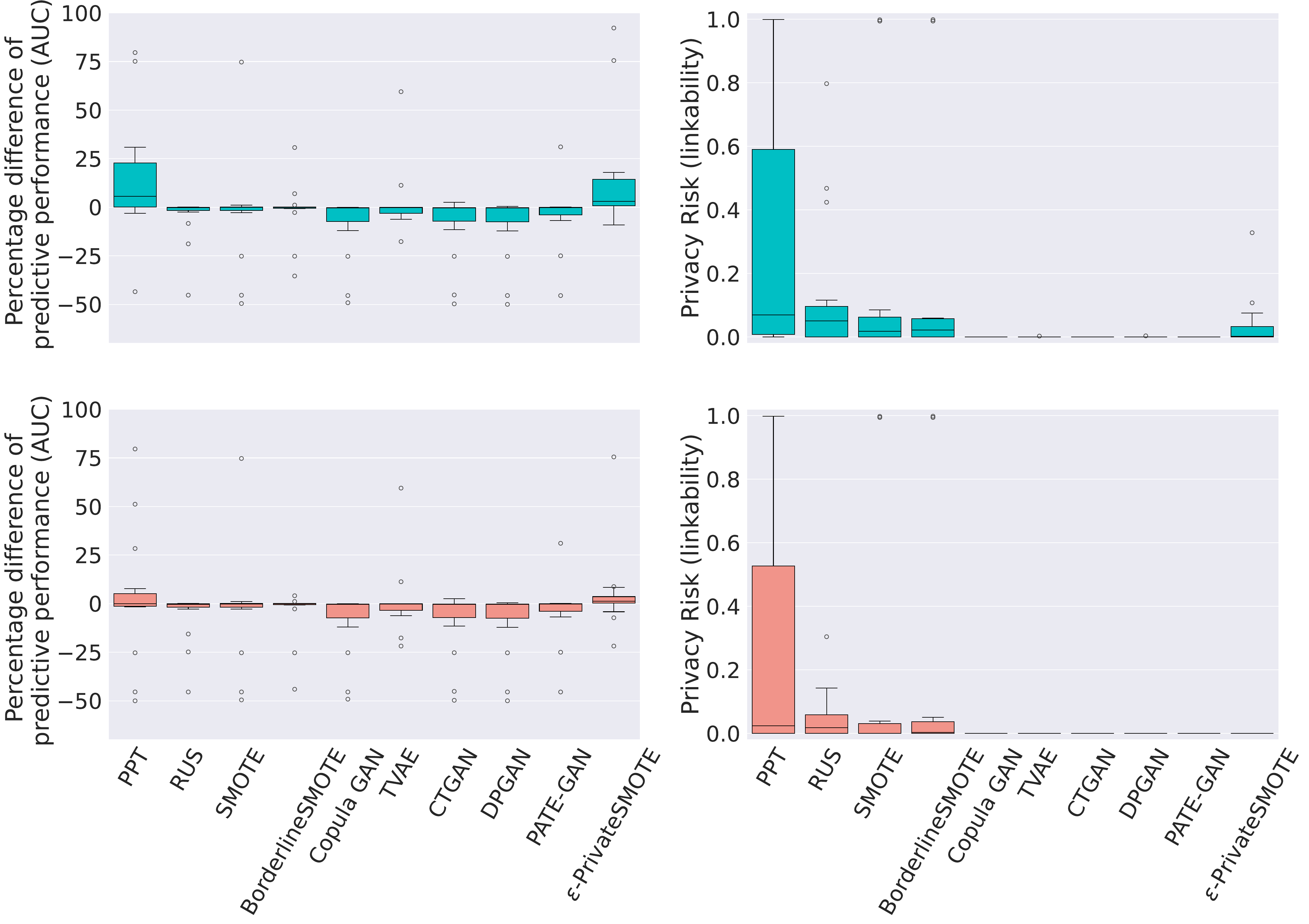}
 \caption{Best predictive performance results and corresponding privacy risk (blue) and best privacy risk results and corresponding predictive performance results for each transformation technique (red).}
 \label{fig:performancerisktogether}
\end{figure}

\paragraph{Optimisation Paths}
In the subsequent analysis, we present a comparative evaluation of $\epsilon$-PrivateSMOTE against state-of-the-art privacy-preservation approaches, explicitly examining the optimisation paths. These paths focus on prioritising a single vector and presenting the corresponding values of the other vector. Thus, 
Figure~\ref{fig:performancerisktogether} demonstrates the effects of selecting the best solutions in predictive performance and the corresponding values for linkability, and the reverse path. Note that PPT refers to traditional techniques.

Focusing on prioritising predictive performance and each approach individually, PPT present a high predictive performance but also a higher linkability risk.  
In our experimental setup, we use lower generalization levels, removal of outliers, and removal of features with too many unique values due to suppression. These transformations allow for obtaining approximate predictive performance values with the original data but could potentiate privacy leaks. 
Regarding sampling methods, the predictive performance is very similar to the original data. Since most of the original data sets are pretty balanced, this may explain the predictive performance results. Also, these three show a certain degree of linkage risk, mainly because they retain instances in their original form.
The scenario is a bit different for deep learning- and differentially private-based solutions. Despite good privacy values, these techniques show a decline in predictive performance. Finally, $\epsilon$-PrivateSMOTE provides a median very similar to PPT with a lower linkability risk, near zero.

Concerning the reverse path, we observe a decrease in linkability risk, especially for $\epsilon$-PrivateSMOTE, which is equivalent to deep learning- and differentially private-based solutions. However, this improvement comes at the cost of predictive performance. This consequence is also demonstrated with PPT. 

In a general overview, $\epsilon$-PrivateSMOTE presents a median with practical equivalence to the baseline while maintaining a reduced linkability risk. Since it does not alter privacy-secure instances, this may help explain its performance. Also, as we produce twice/triple new cases ($N$ parameter), this can worsen the linkage ability because it creates more uncertainty about the cases.

\section{Discussion}\label{sec:discussion}

It is well known that the maximisation of data privacy is at the expense of data utility~\cite{carvalho2022survey}. Thus, it is likely that the higher the privacy risk, the higher the predictive performance. 
However, $\epsilon$-PrivateSMOTE has proven its effectiveness in providing solutions that guarantee lower losses for predictive performance while maintaining lower linkability (Figure~\ref{fig:density}).
Comparing the efficacy in terms of predictive performance and privacy risk of $\epsilon$-PrivateSMOTE with the state-of-the-art approaches (\textbf{RQ1}), the results show that our proposal can produce competitive results with the baseline for both scenarios. $\epsilon$-PrivateSMOTE stands out from the remaining approaches (Figure~\ref{fig:performancerisktogether}), especially compared to PPT, which provides a competitive result in terms of predictive performance. Although PPT preserve non-QI attributes in their original form, $\epsilon$-PrivateSMOTE maintains a similar predictive performance while protecting all attributes with a lower linkability risk.

Concerning the optimisation paths, we observe better linkability results achieved by prioritising privacy. 
This outcome suggests that it is beneficial to prioritise privacy first for $\epsilon$-PrivateSMOTE. On the other hand, the privacy improvements for PPT may not compensate for the loss in predictive performance. For the remaining tested solutions, it is practically indifferent which vector is prioritised (\textbf{RQ2}).

\begin{figure}[ht!]
   \centering
   \includegraphics[width=0.7\linewidth]{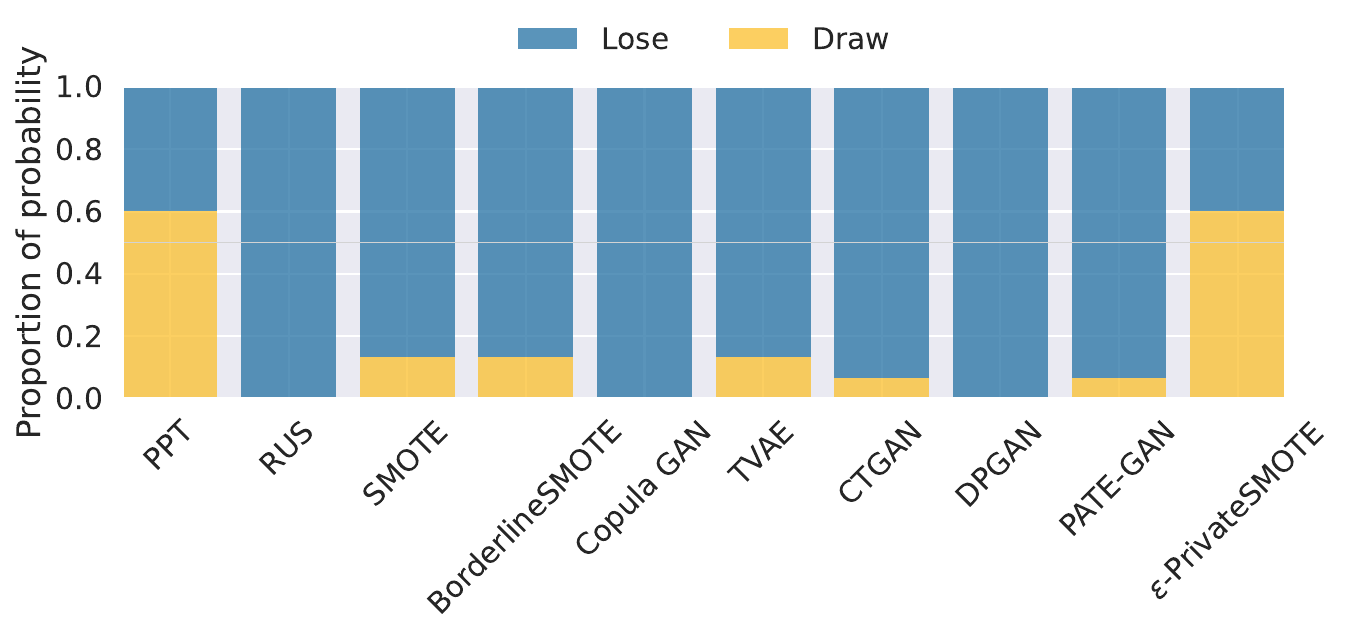}
 \caption{Comparison between the best-estimated hyperparameter configuration per transformation technique and the oracle configuration. Illustrates the proportion of probability for each candidate solution drawing or losing significantly against the oracle according to the Bayes Sign Test for predictive performance.}
 \label{fig:baseline_best}
\end{figure}

To provide a deeper understanding of the difference between $\epsilon$-PrivateSMOTE and the remaining approaches, Figure~\ref{fig:baseline_best} provides a comparison that reports the results of the Bayes Sign Test for predictive performance to infer the statistical significance of the paired differences in the outcome of the approaches. With this aim, we identify the \textit{oracle} model and use it as the baseline regarding privacy prioritisation. 
The oracle identification relies on the assumption that a strategy is able to consistently estimate the best possible model in out-of-sample performance from the explored hyperparameter configurations. Figure~\ref{fig:baseline_best} illustrates the best estimated (in cross-validation settings) model for each transformation technique and the \textit{oracle} for each data set.


Results show that the probability of practical equivalence with the oracle is  60\% for both PPT and $\epsilon$-PrivateSMOTE. Even though privacy is prioritised, these strategies demonstrate the ability to ensure favourable predictive performance results. However, in contrast to the results observed with $\epsilon$-PrivateSMOTE, the higher linkability risk previously identified with the use of PPT is noteworthy.
The remaining transformation techniques present a probability of practical equivalence to the oracle lower than 15\%.

\begin{figure}[ht!]
   \centering
   \scriptsize
   \includegraphics[width=\linewidth]{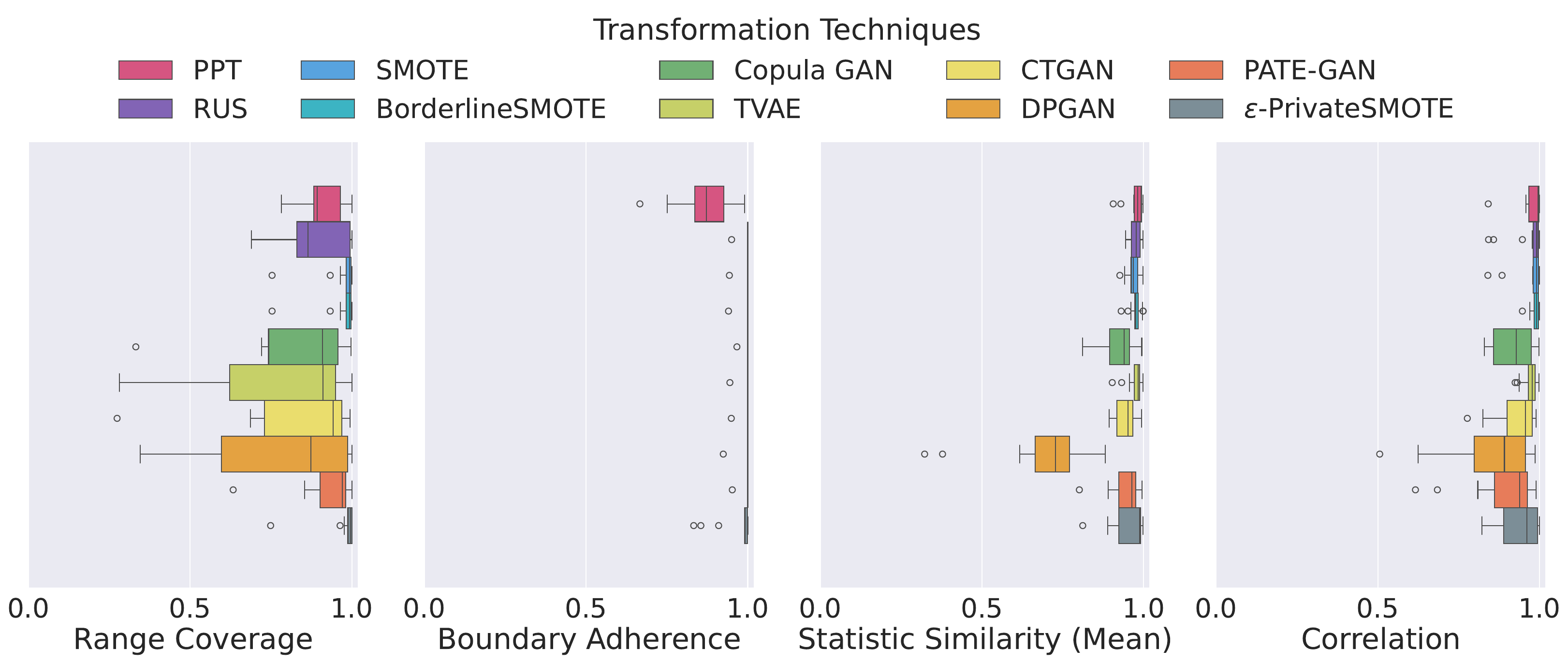}
 \caption{Data utility measures for each transformation technique across best data variants w.r.t privacy optimisation.}
 \label{fig:util}
\end{figure}

Additionally, we provide in Figure~\ref{fig:util} the comparison of all transformation techniques in terms of general data utility for the privacy optimisation path. We transformed the ranges of PPT data variants by selecting the lowest bound to maintain the same data type as the original data.

While PPT present high results for Statistic Similarity (Mean) and Correlation, it holds the worst result for Boundary Adherence, which may also explain the highest linkability risk. Concerning deep learning- and differentially private-based solutions, the Boundary Adherence is high but the remaining measures present a higher variability, which is reflected in the predictive performance results (Figure~\ref{fig:performancerisktogether}). $\epsilon$-PrivateSMOTE also shows some variability for Statistic Similarity (Mean) and Correlation, probably due to the oversampling of the highest-risk cases. In summary, SMOTE, BorderlineSMOTE and $\epsilon$-PrivateSMOTE present a median very close to the maximum for all data utility measures. However, only $\epsilon$-PrivateSMOTE provides a good trade-off between linkability and predictive performance.

Although $\epsilon$-PrivateSMOTE can provide competitive results, the biggest advantage is its simplicity (\textbf{RQ3}). We compare deep learning- and differentially private-based solutions with $\epsilon$-PrivateSMOTE in terms of data variant generation. As shown in Table~\ref{tab:costs}, $\epsilon$-PrivateSMOTE can generate a large number of data variants much faster while consuming much fewer resources than the compared solutions. We do not present the computational cost results for the PPT as the parameters were handset. Sampling methods are omitted due to their linkability risk, making them non-competitive.

\begin{table}[!ht]
\begin{center}
    \scriptsize
    \begin{adjustbox}{max width=0.7\linewidth}
\begin{tabular}{@{}l|cccccc@{}}
\toprule
\textbf{Technique}       & \textbf{\begin{tabular}[c]{@{}c@{}}Nr. of generated \\ data variants\end{tabular}} &  \textbf{\begin{tabular}[c]{@{}c@{}}Elapsed \\ Time (min)\end{tabular}} & \textbf{\begin{tabular}[c]{@{}c@{}}CPU \\ (\%) \end{tabular}} & \textbf{\begin{tabular}[c]{@{}c@{}}GPU \\ (\%) \end{tabular}} & \textbf{\begin{tabular}[c]{@{}c@{}}RAM \\ (\%) \end{tabular}} & \textbf{\begin{tabular}[c]{@{}c@{}}Time \textbackslash \\ Variant\end{tabular}} \\ \midrule
\textit{Copula GAN}      & 60                                                                                & 207,60                                                                             & 10,43     & 17,60 & 37,42 & 3,46                                                                     \\
\textit{TVAE}            & 60                                                                                & 125,26                                                                              & 13,09     & 16,76 & 37,42 &2,08                                                                     \\
\textit{CTGAN}           & 60                                                                                & 206,94                                                                             & 10,59      & 41,66  &  37,44 & 3,44                                                                     \\
\textit{DPGAN}           & 240                                                                                & 857,37                                                                             & 8,19     & 23,07  &47,34 & 3,57                                                                     \\
\textit{PATE-GAN}           & 240                                                                                & 355,33                                                                            & 14,59     & 9,96  &47,42 & 1,48                                                                     \\

\textit{$\epsilon$-PrivateSMOTE}    & 3375   & 553,28                                                           & 22,41     & 0  & 22,41& 0,16                                                                      \\ \bottomrule
\end{tabular}
\end{adjustbox}  
\caption{Comparison of computational costs between deep learning- and differentially private-based models and $\epsilon$-PrivateSMOTE.}
\label{tab:costs}
\end{center}
\end{table}

Table~\ref{tab:costs} provides the number of data variants generated for each approach, considering the number of QI set and the number of parameters of the transformation techniques. For example, PATE-GAN produces 16 data variants for each original data set, resulting in a total of 240 data variants. 
It is noticeable that despite $\epsilon$-PrivateSMOTE generating more than 14 times the number of variants, it demonstrates efficiency by being approximately 9 times faster than PATE-GAN and using fewer resources. We experimented with GPU utilisation; however, our proposed approach requires more time to generate a data variant on the GPU due to a portion of operations running on the CPU and others on the GPU. The operations between the CPU and GPU introduce additional latency. Therefore, we opted to implement $\epsilon$-PrivateSMOTE on the CPU for improved efficiency.
Such results were conducted on a Linux-based system with the following specifications: kernel version 5.15.0-91-generic, 2.85GHz 24-Core AMD EPYC Processor, 256GiB RAM and a GeForce RTX 3090.

To conclude, $\epsilon$-PrivateSMOTE emerges as a robust method that provides similar or better privacy guarantees than traditional and generative methods while optimising predictive performance, and exhibiting similar statistical properties with original data. Most importantly, it significantly reduces computation time and operates with reduced resource requirements when generating new data variants.
However, we still face some limitations, especially regarding potential inference-based attacks.  
A further concern is the obfuscation of outliers. A data set generated with $\epsilon$-PrivateSMOTE may still contain outliers, which can have significant privacy implications. Also, our solution focuses on tabular data and cannot handle other types of data such as images.
All of these potential issues will be investigated in future work.

\section{Conclusions}\label{sec:conclusion}

In this paper, we propose $\epsilon$-PrivateSMOTE, a new strategy for privacy-preserving data sharing that combines data synthetisation via noise-induced interpolation with Differential Privacy to address the highest-risk cases concerning $k$-anonymity.  
Experimental results show that: \textit{i)} $\epsilon$-PrivateSMOTE is capable of providing competitive results concerning predictive performance; \textit{ii)} it provides lower linkability risk, especially when privacy is the priority; and, \textit{iii)}  
$\epsilon$-PrivateSMOTE proved to be a more resource-efficient solution than deep learning- and differentially private-based approaches, improving time complexity by a factor of 9. Moreover, it performs well on lower dimensional data sets and does not require learning all real data.






\section*{Declarations}

\begin{itemize}
    \item Funding: Not applicable.
    \item Conflicts of interest/Competing interests: I declare that the authors have no competing interests as defined by Springer, or other interests that might be perceived to influence the results and/or discussion reported in this paper.
    \item Ethics approval: Not applicable.
    \item Consent to participate: Not applicable.
    \item Consent for publication: all of the material is owned by the authors and/or no permissions are required.
    \item Availability of data and material: Data is provided within the manuscript.
    \item Code availability: The code of the proposed method is available at https://github.com/tmcarvalho/privateSMOTE.
    \item Authors' contributions: Tânia Carvalho and Nuno Moniz wrote the main manuscript. All authors reviewed the manuscript.
\end{itemize}


\bibliography{sn-bibliography}

\end{document}